\documentclass{article}

\usepackage[final]{neurips_2022}

\usepackage[utf8]{inputenc} 
\usepackage[T1]{fontenc}    
\usepackage{url}            
\usepackage{booktabs}       
\usepackage{amsfonts}       
\usepackage{nicefrac}       
\usepackage{microtype}      
\usepackage{lineno}
\usepackage{natbib}
\usepackage{graphicx}
\usepackage{hyperref}       
\usepackage{xcolor}         

\title{Decentralized Technologies for AI Hubs}

\author{%
  Richard Blythman\\
  Algovera\\
  Ireland \\
  \texttt{richard@algovera.ai} \\
  \And
  Mohamed Arshath\\
  Algovera\\
  Malaysia \\
  \texttt{arshy@algovera.ai} \\
  \And
   Salvatore Vivona\\
  Algovera\\
  Canada \\
  \texttt{sal@algovera.ai} \\
  \And
   Jakub Smékal\\
  Algovera\\
  United Kingdom \\
  \texttt{jakub@algovera.ai} \\
  \And
  Hithesh Shaji\\
  Algovera\\
  Ireland \\
  \texttt{hithesh@algovera.ai} \\
}

\begin{document}

\maketitle

\begin{abstract}

AI requires heavy amounts of storage and compute with assets that are commonly stored in AI Hubs. AI Hubs have contributed significantly to the democratization of AI. However, existing implementations are associated with certain benefits and limitations that stem from the underlying infrastructure and governance systems with which they are built. These limitations include high costs, lack of monetization and reward, lack of control and difficulty of reproducibility. In the current work, we explore the potential of decentralized technologies - such as Web3 wallets, peer-to-peer marketplaces, storage and compute, and DAOs - to address some of these issues. We suggest that these infrastructural components can be used in combination in the design and construction of decentralized AI Hubs. 

\end{abstract}

\section{Introduction}

The field of deep learning is powered by assets such as datasets, models and software, which require heavy amounts of storage and compute \cite{schwartz2020green}. As a result, data scientists are regular users of AI Hubs such as GitHub, Kaggle, HuggingFace Hub and ActiveLoop Hub to provide a place for assets to be stored, shared and further developed. AI Hubs have been a significant factor in democratizatizing access to state-of-the-art pretrained models \cite{wolf2019huggingface} and the contribution of open source to the field of AI \cite{langenkamp2022open}. 

At the same time, existing AI Hubs make certain trade-offs that arise from their underlying technologies and governance structures \cite{marketplace}. Today's AI Hubs tend to rely exclusively on centralised cloud services such as AWS, GCP, and Azure, and the high expense of these services is often passed on to the user. Furthermore, while the assets themselves may be open source, the platform itself is typically closed source and governed by a centralized entity. The platform ultimately controls the accessibility of uploaded assets, and monetizes the network effects of user contributions without sharing in the rewards. Finally, the assets within AI Hubs are often isolated from compute and not well integrated with workflows, making reproducibility difficult. 

Decentralized technologies such as peer-to-peer storage, compute and marketplaces, machine learning frameworks and decentralized autonomous organizations (DAOs) present opportunities for tackling the above challenges. We explore the benefits offered by these technologies to address some of the above issues within decentralized AI hubs, which offer an alternative value proposition compared with existing solutions. 

\section{Limitations of Existing AI Hubs}

An AI Hub is a platform that allows data scientists, engineers and other stakeholders to share and discover AI assets such as  datasets, models, apps, notebooks, pipelines and other software. AI Hubs require different features and infrastructure such as storage and compute, and may also provide organisational tools on top. 

There are a number of existing hubs such as GitHub, Kaggle, HuggingFace (HF) and ActiveLoop with different features as summarized in Table \ref{tab:hubs}. GitHub, owned by Microsoft, is a popular platform for storing software assets. HF achieved success by standardising the code for architectural components and providing a unified API for the popular transformer model for natural language processing (NLP) use cases \cite{wolf2019huggingface}. As well as storing code, HF Hub provides storage for datasets and models, and compute for demos and inference. ActiveLoop Hub focuses on efficient cloud streaming of datasets for deep learning. Replit is an online integrated development environment (IDE). Unlike GitHub and HuggingFace, where modifying assets requires a separate IDE and command line, Replit users can interact with code and source control for their project through a web-based graphical user interface. Replit provides a shared compute engine that provides collaborative coding similar to Google Docs, where code can be run and displayed to multiple users. However, GPU support has not yet been released. Furthermore, file storage is limited to 0.5 GB for free users and 5 GB for paid users, which is too small for most ML assets. In general, existing AI Hubs are built using centralized infrastructure, which have certain benefits and limitations. Replit has other features such as AI-assisted tools for software development, such as co-pilot and live chat and in-line threads for discussions around code by users. 

\begin{table}[!htp]\centering
\caption{Existing AI Hubs}\label{tab:hubs}
\scriptsize
\begin{tabular}{lrrrrr}\toprule
\textbf{Existing AI Hub} &\textbf{GitHub} &\textbf{Kaggle} &\textbf{Huggingface} &\textbf{ActiveLoop} &\textbf{Replit} \\\midrule
\textbf{Launch} & 2008 & 2010 & 2016& 2018 & 2016 \\
\textbf{Users} & SWEs & Data Scientists & Data Scientists & Data Scientists & SWEs \\
\textbf{Monetization} & No & Prizes & No & No & No \\
\textbf{Storage/Asset} & Code & Code, Datasets, Notebooks & Code, Datasets, Models, Apps & Datasets & Code \\
\textbf{Compute/Hosting} & No & GPU (Notebooks) & GPU (Inference, Apps) & No & CPU \\
\textbf{Cloud Infrastructure} & Centralized & Centralized & Centralized & Centralized & Centralized \\
\textbf{Governance} & Centralized & Centralized & Centralized & Centralized & Centralized \\
\bottomrule
\end{tabular}
\end{table}

\subsection{High Storage and Compute Costs} \label{s:cost}

The computational cost of AI research is increasing exponentially, creating to higher barriers to entry for participants \cite{schwartz2020green}. As a result, cloud services, such as storage and compute, are a significant expense for AI startups. Currently, three companies make up approximately two thirds of the market share of cloud service \cite{cloudmarket}. More than half of Amazon’s profits has come from Amazon Web Services, and 20\% of AWS customers deliver 80\% of revenue with the widest margins come from small and medium-sized customers \cite{amazon}. Popular AI Hubs like GitHub, HuggingFace, ActiveLoop and Replit rely exclusively on centralised cloud platforms.

\subsection{Lack of Monetization and Reward} \label{s:Ownership}

There are few online platforms where data scientists can perform paid work independently \cite{marketplace}. Within existing AI Hubs, money only travels from the user to the platform itself. While AI Hubs like HuggingFace do offer free services and contribute to open source development, they also charge users for premium services that are not open source. In contrast, all contributions by users must be open source, with no ability to offer paid services. 

Open source tools and libraries are widely used by commercial platforms and products within software development and AI \cite{langenkamp2022open}, although the contributions are not typically rewarded. Platforms invite assets to be uploaded by users, but do not share any generated revenue or platform ownership with users, even when directly monetizing their contributions. For example, GitHub Copilot is a commercial product for code generation that uses a model trained on user-contributed code. HuggingFace's paid inference API can be used to accelerate the deployment of user-contributed models. 

\subsection{Lack of Control} \label{s:Control}

Generally, software developers and data scientists do not have full control and autonomy with their creations on centralized platforms. In one case, GitHub reverted malicious changes (and suspended the account) of a developer to their own popular open source library, raising questions around the rights of developers to do what they wish with their code \cite{developer}. In the field of AI, there has been an ongoing discussion on whether open sourcing disruptive models should be commonplace, since there is the potential for harm and bias. For example, AlphaFold can be used for discovery of novel toxic molecules. Language models can be trained on abusive content and used by online bots. Large models that are trained on the corpus of internet data reproduce bias within generated text and images. As a result, platforms like HuggingFace have come under pressure to gate or remove access to models.  On the other hand, it can be argued that open sourcing the model puts the technology in the hands of more people that can study and solve issues around safety and bias. In other words, there is an orthogonal risk involved with centralization of AI in the hands of a few. Keeping models closed source effectively turns large tech companies into gatekeepers, who may not always be relied upon to adjudicate on disputes in an unbiased manner. 

Finally, it is difficult for owners to manage fine-grained access to assets. Traditional access tokens like OAuth 2.0 \cite{hardt2012oauth} and API keys for datasets and models can be widely shared, and licenses for datasets and software are often broken. While possible to keep repositories private, this is often a paid feature and the encryption key is held by the platform rather than the user.

\subsection{Difficult to Reproduce} \label{s:Reproduce}

The limitations of existing hubs such as GitHub for AI can make reproduciblity more difficult. For example, academic papers often contain links to AI Hubs for the purpose of reproducibility. This may include code on GitHub, and datasets and model weights stored on the cloud. Nonetheless, reproducing experiments is difficult and can require many steps such as downloading datasets, running processing scripts and installing environments. This issue results from a variety of factors such as the lack of standardisation and interoperability of in the format of assets (such as dataset and code), and the decoupling of assets from compute environments and infrastructure needed to operate on them. Some of these issues can be resolved by using containers and notebooks to replicate environments and bring compute to code. HuggingFace Hub uses Gradio and Streamlit apps. Replit integrates code repositories with compute environments, but has limited storage for assets such as datasets and model weights. 




\section{Decentralized Technologies for AI Hubs}

\begin{figure}[htb]
    \centering
    \includegraphics[width=0.8\textwidth]{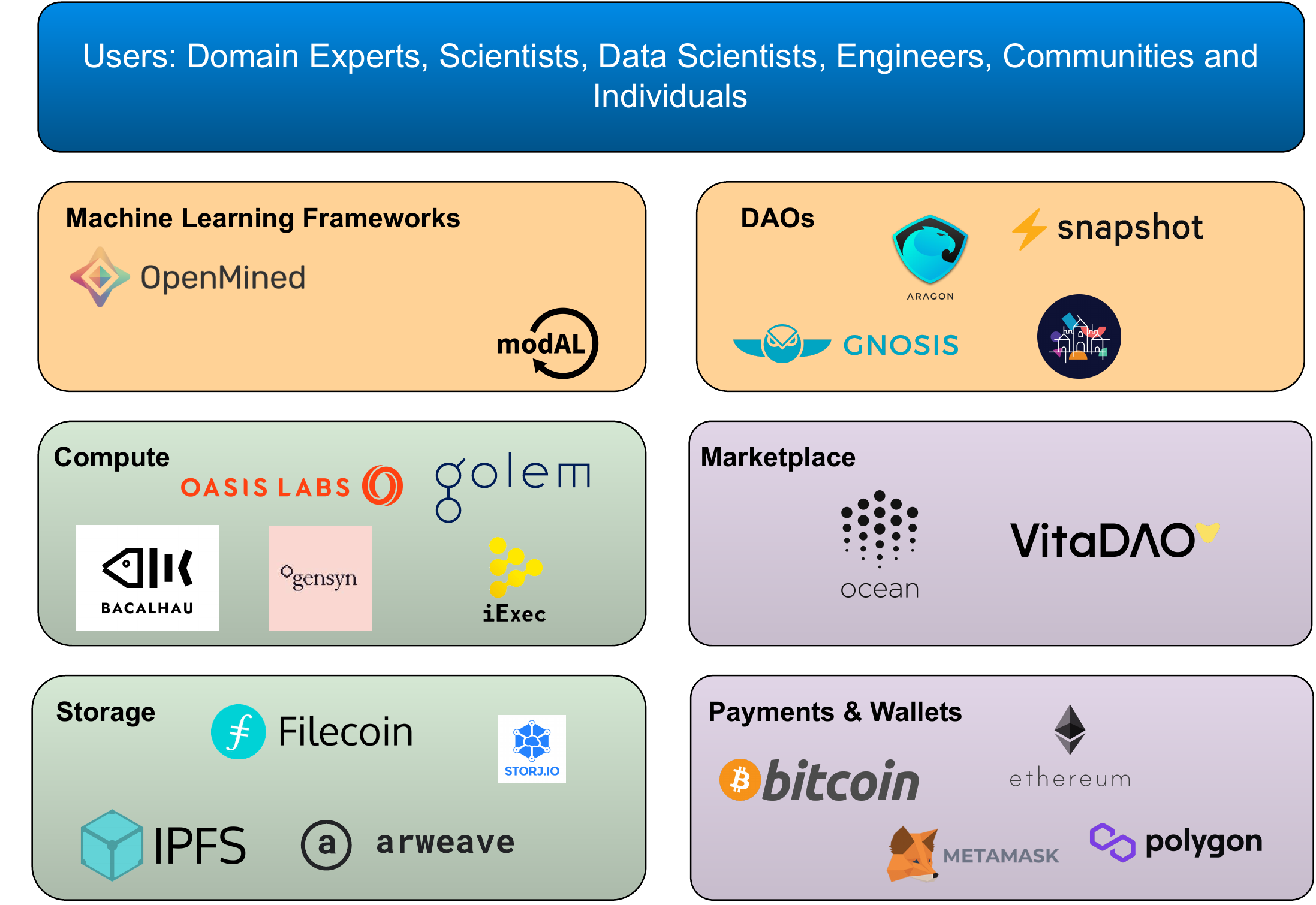}
    \caption{The decentralized AI stack (or Web3 AI stack), consisting of decentralized technologies that offer opportunities for decentralized AI Hubs. The users of decentralized AI Hubs are the many stakeholders required for undertake successful projects.}
    \label{fig:metahub}
\end{figure}

Decentralized technologies - such as Web3 payments, wallets, marketplaces, storage and compute, learning frameworks and DAOs - have the potential to alleviate some of the limitations of existing AI Hubs discussed above. Examples of projects working on these individual projects are shown in Figure \ref{fig:metahub}.

\subsection{Payments} 

There are few options for AI workers to monetize their creations and rewards generated by their contributions are often not shared, as discussed in Section \ref{s:Ownership}. We believe that building in mechanisms for monetization and ownership by users would create a healthier and more sustainable ecosystem and economy. This can be achieved using cryptocurrencies (such as Bitcoin, Ethereum, Polygon, Ocean and Filecoin) and stablecoins (such as DAI or USDC), which can be used for micro- and streaming payments to stakeholders such as data scientists, data providers and compute providers with low transaction fees. Thus, integrated payments offer many opportunities for use with machine learning frameworks such as active learning and data crowdsourcing.


\subsection{Web3 Wallets} 

As discussed in Section \ref{s:Control}, data scientists typically do not have control of what they create online. Even if a repository containing assets is private, the platform holds the private keys. A Web3 wallet can be used to put the user in control of their private keys. The word wallet tends to have financial connotations. However, wallets are often used in the real world as a place where you hold ownership and identity documents (such as a driver’s license). Similarly, Web3 wallets can be used for ownership and identity in the digital world. Wallets are interoperable in the sense that you can use the same wallet to signify ownership of assets across many different protocols. Web3 wallets include software wallets (such as MetaMask\footnote{\url{https://metamask.io/}}) and hardware wallets (such as Trezor\footnote{\url{https://trezor.io/}}).

\subsection{Marketplace} Traditional AI Hubs and marketplaces are typically operated by a centralized entity serving as a middle man. In the ideal scenario, the operator provides services in exchange for transaction fees and acts as a mediator for conflict resolution between users. However, centralized hubs and marketplaces also have the power to capture an outsized proportion of the value generated in a market economy as network effects grow. This contributes to the issues discussed in Sections \ref{s:Ownership} and \ref{s:Control}. 

Using decentralized marketplaces protocols for tracking publication, ownership of (and access to) assets has the potential to mitigate these risks. All operations are stored on an immutable public distributed ledger such that provenance can be tracked. For AI use cases, assets can include datasets, models, algorithms, apps, notebooks and manuscripts. Examples of decentralized marketplaces include Ocean Protocol \cite{protocol2021tools} and VitaDAO \cite{golatowhitepaper}. These protocols use non-fungible tokens (NFTs) to represent ownership of the underlying intellectual property (IP), and fungible tokens to represent access rights to assets under different types of licenses. The details of published assets (and associated metadata) are encrypted and stored on-chain, along with access control parameters. A decentralized identifier (DID) is issued to represent the asset’s decentralized digital identity, and a DID Document (DDO) is used to include additional information relevant to the asset. 

Access gated by tokens on a blockchain has advantages compared to traditional access token like OAuth 2.0, by solving the "double spend problem". They act as access tokens that can only be used by one individual or for a period of time. If a user receives a token on a blockchain, the user can still share it with someone else but this means the original user will no longer have access. This facilitates more fine-grained access control by owners. 

\subsection{Storage} 

While details about the assets are stored on-chain with decentralized marketplaces, the data associated with the asset are often too large to store on chain. As discussed in Section \ref{s:cost}, storage on centralized cloud providers is expensive. Furthermore, these services are less robust and more prone to censorship (see Section \ref{s:Control}). Popular dataset and model hubs like HuggingFace and ActiveLoop Hub rely on centralised cloud platforms. 


Decentralised protocols for storage have the potential to vastly reduce the costs incurred by data scientists for storing raw and processed versions of datasets and model weights. This makes  it possible to download files from multiple locations that aren't managed by a single organization.  The interplanetary file system (IPFS) \cite{ipfs} is a   peer-to-peer  protocol for storing and accessing data in a permissionless and censorship-resistant way. IPFS clusters enable data orchestration across swarms of IPFS peers by allocating, replicating, and tracking assets. Another important feature that IPFS offers is the ability to verify the validity of assets using Content Addressable Identifiers (CIDs), based on the content's cryptographic hash. 



\subsection{Compute} 

Access to compute is a necessity for AI projects, and the provision of services by a handful of centralized companies has resulted in inflated costs (see Section \ref{s:cost}). At the same time, the experiments and results of AI studies are often difficult to reproduce, as discussed in Section \ref{s:Reproduce}. While less mature than peer-to-peer storage solutions, decentralized protocols for providing compute resources aim to reduce the barrier-to-entry for compute providers and remove the centralised overheads on scaling \cite{fieldingwhitepaper}. This provides more options for end consumers, resulting in reduced cost. Compute can be run where the data is stored - called Compute over Data (CoD) by the Bacalhau project\footnote{\url{https://github.com/filecoin-project/bacalhau}}, or Compute to Data (C2D) by Ocean Protocol - rather than transporting data to the location of the compute which is expensive. In this setting, decentralized compute infrastructure presents many opportunities for integration with privacy-preserving machine learning. Running compute jobs in a trustless setting requires verification that it was carried out correctly, which can be difficult for non-deterministic compute such as deep learning. Gensyn\footnote{\url{https://www.gensyn.ai/}} have developed a novel system for providing proof under this condition.



\subsection{Machine Learning Frameworks} 

Decentralizing infrastructure for storage and compute, and integrating payments has the potential to open up new use cases of AI. This require advancements in decentralized frameworks for machine learning. For example, privacy-preserving machine learning (PPML) - through libraries such as Openmined\footnote{\url{https://www.openmined.org/}} - has the potential to unlock learning on private data such as health records and user data. Integrated payments can be used with active learning frameworks with libraries (such as modAL \cite{danka2018modal}) and tools for crowdsourcing human intelligence (such as Turkit \cite{little2010turkit}). 

\subsection{DAOs}  

Decentralized autonomous organizations (DAOs) are systems that allow communities to coordinate and take part in self-governance, as determined by a set of self-executing rules on a blockchain \cite{hassan2021decentralized}. DAOs have previously been suggested as governance structures for digital data trusts \cite{nabben2021decentralised}. We suggest that DAOs can be used to (i) govern assets within AI Hubs, and (ii) to create decentralized AI Hubs that are governed by communities rather than single entities. Tools for governing assets within DAOs include multisignature wallets (such as Gnosis\footnote{\url{https://gnosis-safe.io/}}) and profit-sharing mechanisms (such as Superfluid\footnote{\url{https://www.superfluid.finance/}}). Multisignature wallets provide functionality for sharing ownership and control of assets with multiple individuals in teams in a trustless manner, while profit-sharing mechanisms can be used to distribute the revenue generated by assets. Tools for governing the infrastructure of AI Hubs include decentralized voting systems (such as Snapshot\footnote{\url{https://snapshot.org/}}).

\section{Conclusion}

In this work, we reviewed the trade-offs made by existing AI Hubs, and explored the ability of a collection of decentralized technologies to mitigate some of their limitations. Decentralized AI Hubs have the potential to reduce the barrier-to-entry for cloud infrastructure, increase monetization opportunities for independent AI teams, put ownership and control in the hands of creators, and improve reproducibility of research. 


\bibliographystyle{plain}
\bibliography{main.bib}%

\small

\end{document}